\documentclass[letterpaper, 10 pt, conference]{ieeeconf}  

\IEEEoverridecommandlockouts                              
                                                          
\usepackage{amsmath} 
\usepackage{amssymb}  
\usepackage{dblfloatfix}    
\usepackage{graphicx}
\usepackage{epstopdf}
\usepackage{amsmath}
\usepackage{mathtools}
\usepackage{dsfont}
\usepackage{tikz}
\usepackage{siunitx}
\usepackage{xcolor}
\usepackage[bottom]{footmisc} 
\usepackage{dsfont}
\usepackage{url}
\usepackage{booktabs} 
\usepackage{multicol}
\usepackage{graphicx}

\usepackage{hyperref}

\usepackage{balance}    
\let\proof\relax
\let\endproof\relax
\usepackage{amsthm}
\usepackage{subfig}
\usepackage{adjustbox}

\usepackage[font=footnotesize,labelfont=bf]{caption}
\usepackage{balance}    
\usepackage{xcolor}
\usepackage{tcolorbox}
\usepackage{dsfont}
\usepackage{algorithm}
\usepackage[noend]{algorithmic}
\floatname{algorithm}{\footnotesize Algorithm}
\usepackage{lipsum} 

\newfloat{usecase}{luc}{Use case}

\usepackage{dsfont}
\usepackage{booktabs} 

\newcommand{\beq}{\begin{equation}}
\newcommand{\eeq}{\end{equation}}

\newcommand{\Nobs}{N}

\newcommand{\Xfree}{\mathcal{X}^a_{\textrm{free}}}
\newcommand{\Xobs}{\mathcal{X}^a_{\textrm{obs}}}

\newcommand{\Par}{\texttt{P}}
\newcommand{\Obs}{\texttt{O}}
\newcommand{\Deltax}{\Delta x}
\newcommand{\Deltas}{\Delta s}

\newcommand{\eqdef}{\mathrel{\mathop:}=}
\newcommand{\Noa}{A}

\newcommand{\astfootnote}[1]{%
\let\oldthefootnote=\thefootnote%
\setcounter{footnote}{0}%
\renewcommand{\thefootnote}{\fnsymbol{footnote}}%
\footnote{#1}%
\let\thefootnote=\oldthefootnote%
}

\def\mathcolor#1#{\@mathcolor{#1}}
\def\@mathcolor#1#2#3{%
  \protect\leavevmode
  \begingroup
    \color#1{#2}#3%
  \endgroup
}

\IEEEoverridecommandlockouts

\usepackage[noadjust]{cite} 
\usepackage{amsmath,stmaryrd,graphicx}

\newtheoremstyle{boldStyle}
  {\topsep}
  {\topsep}
  {\itshape}
  {0pt}
  {\bfseries}
  {.}
  { }
  {\thmname{#1}\thmnumber{ #2}\thmnote{ (#3)}}

\newtheoremstyle{italicStyle}
  {\topsep}
  {\topsep}
  {}
  {0pt}
  {\bfseries}
  {.}
  { }
  {\thmname{#1}\thmnumber{ #2}\thmnote{ (#3)}}

\theoremstyle{boldStyle}

\newtheorem{proposition}{Proposition}

\newcommand{\Prob}{\mathbb{P}}

\theoremstyle{italicStyle}

\makeatletter
\newcommand{\fixed@sra}{$\vrule height 2\fontdimen22\textfont2 width 0pt\shortrightarrow$}
\newcommand{\shortarrow}[1]{%
  \mathrel{\text{\rotatebox[origin=c]{\numexpr#1*45}{\fixed@sra}}}
}
\makeatother

\title{\LARGE \bf 
Mixed Observable RRT: Multi-Agent Mission-Planning \\in Partially Observable Environments}

\author{Kasper Johansson, Ugo Rosolia, Wyatt Ubellacker, Andrew Singletary, and Aaron D. Ames}%

\author{%
  Kasper Johansson%
  \rlap{\textsuperscript{1}},
  Ugo Rosolia%
  \rlap{\textsuperscript{2}},
  Wyatt Ubellacker%
  \rlap{\textsuperscript{2}},
  Andrew Singletary%
  \rlap{\textsuperscript{2}},
  and Aaron D. Ames%
  \rlap{\textsuperscript{2}}
}

\begin{document}

\maketitle
\global\csname @topnum\endcsname 0
\global\csname @botnum\endcsname 0

\footnotetext[1]{\scriptsize{Stanford University, Stanford, USA. \texttt{kasperjo@stanford.edu}. This work was done while Kasper was at KTH Royal Institute of Technology, Stockholm, Sweden.}}
\footnotetext[2]{\scriptsize{California Institute of Technology, Pasadena, USA. \texttt{\{urosolia, wubellac, asinglet, ames\}@caltech.edu}}}

\begin{abstract}
This paper considers centralized mission-planning for a heterogeneous multi-agent system with the aim of locating a hidden target. We propose a mixed observable setting, consisting of a fully observable state-space and a partially observable environment, using a hidden Markov model. First, we construct rapidly exploring random trees (RRTs) to introduce the mixed observable RRT for finding plausible mission plans giving way-points for each agent. Leveraging this construction, we present a path-selection strategy based on a dynamic programming approach, which accounts for the uncertainty from partial observations and minimizes the expected cost. Finally, we combine the high-level plan with model predictive control algorithms to evaluate the approach on an experimental setup consisting of a quadruped robot and a drone. It is shown that agents are able to make intelligent decisions to explore the area efficiently and locate the target through collaborative actions. 
\end{abstract}

\section{Introduction} \label{sec:intro}
\begin{figure}[t]
    \centering
    \includegraphics[width=\linewidth, trim={30 110 10 120},clip]{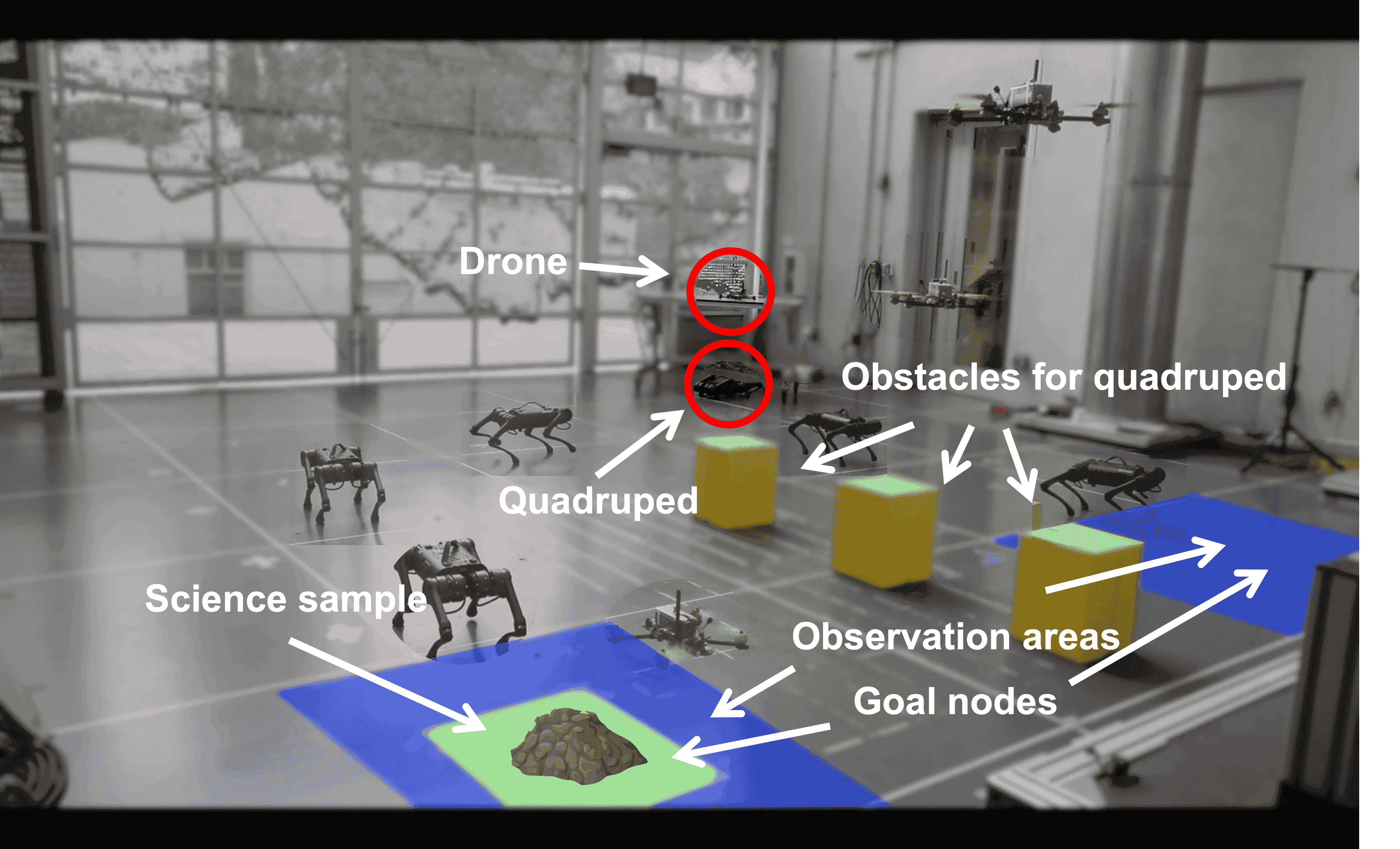}
    \caption{A quadruped and a drone are initiated inside the red circles. They locate a science sample by sharing observations from the environment.}
    \label{fig:experimentalSetup}
\end{figure}
Planning under uncertainty is important for autonomous systems. Similar to human decision-making based on observations---for instance, when driving or playing sports---autonomous robotic systems must be able to act based on observations from the environment. Exploration tasks are one type of problem where robots have partial knowledge about the environment, complemented with observations along the mission. There are several important instances of exploration tasks. One motivating example is the Mars explorations task, where the objective is to allow robots to act autonomous and gather information on Mars~\cite{marsExploration}. However, there seems to exist no satisfying solution to such exploration when multiple robots are to cooperate and act simultaneously to locate a hidden target in a partially known environment.

\subsection{Contribution}
In this work we introduce a novel approach to path-planning in partially observable environments for multi-agent systems. Agents act simultaneously to locate a target, as illustrated in Figure~\ref{fig:experimentalSetup}. We leverage rapidly exploring random trees (RRTs) to generate a sample of event-based plans, where a plan consists of a sequence of way-points that changes as a function of the realized observations agents make. The best plan is found by minimizing an expected cost, formulated as a summation over nodes in trajectory trees, as proposed in~\cite{rosolia2021mixedobservable}. Lastly, to allow a multi-agent system to follow the high-level plan we leverage local model predictive control (MPC) algorithms. 

Our contribution is threefold. First, we model a cooperative multi-agent exploration mission using a mixed observable setting. We modify the standard RRT algorithm to generate a sample of plans in the discrete partially observable environment, defining the \textit{mixed observable RRT (MORRT)}. Secondly, we illustrate the advantage of the MORRT over standard RRTs in uncertain environments, and apply the algorithm on a system of five agents. Finally, we demonstrate our approach experimentally on hardware by integrating the MORRT plan, consisting of way-points for a heterogeneous multi-agent system, with local MPC algorithms.

The paper is outlined as follows. Section~\ref{sec:intro} reviews previous works. In Section~\ref{sec:problemFormulation} we formulate the problem and describe the mixed observable setting. Section~\ref{sec:MORRT_section} details our main contribution, the sample-based search algorithm for mixed observable settings, denoted the MORRT. In Section~\ref{sec:results} we report results from simulations and a hardware experiment with a quadruped robot and a drone, as well as illustrate the advantage over standard RRTs in uncertain environments. Finally, we summarize our findings in Section~\ref{sec:conslusion}.

\subsection{Related works}

\textbf{Planning.} The literature on environment mappings for navigation and planning is rich~\cite{marsExploration, lakemeyer2003, thrun2005probabilistic, stachniss2009}. Exploration can be done using a policy that maps the system's state to a control action~\cite{rosolia2021mixedobservable}. The computational burden associated with planning over policies has been extensively studied~\cite{GOULART2006523, CHISCI20011019, MAYNE2005219, YU2013194, fleming2015, wang2019, liniger2017}, and can be reduced when a system's state can be perfectly measured~\cite{rosolia2021mixedobservable, tal2004, GOULART2006523, wang2019}. Tube MPC strategies involve another type of feedback policies~\cite{fleming2015, YU2013194, MAYNE2005219, CHISCI20011019}. Strategies where only partial state knowledge is available have been studied in~\cite{MAYNE20061217, alvarado2007, CANNON2012536}. In~\cite{rosolia2021mixedobservable} the authors consider multi-modal measurement noise by planning over a tree of trajectories where each branch is associated with a unique set of observations. We propose a novel way to tackle planning problems under partial observations, based on RRTs.

\textbf{Rapidly exploring random trees.} A popular tool for path-planning is the RRT, which is a sampling-based search algorithm~\cite{LaValle2001}. RRTs have been leveraged for path-planning in various problems, like navigating among moving obstacles~\cite{fulgenzi2008}, kinodynamic motion-planning~\cite{jongwoo2003}, and probabilistically robust path-planning~\cite{kothari2013}, among many others. The RRT$^{*}$ algorithm is an extension of the RRT~\cite{RRT_extend}, and it has been shown that RRT$^{*}$ converges to the optimal path~\cite{RRTstarKaraman2}. Although RRTs have been shown to perform well in certain tasks, they lack in that they do not allow for information gathering in the form of observations. In this work, we extend the RRT to allow for drastic changes in the path when observations are made and the target belief changes.

\textbf{Coordinated multi-robot exploration.} Coordinated exploration for multi-robot systems has been extensively studied, e.g.,~\cite{collaborative_exploration_Burgard2000, coordinated_segmentation_Burgard2008, coordinated_marsupial_teams_Burgard2010, coordinated_exploration_Burgard2005, cooperating_cleaning_robots_Jager2002, cooperative_sweeping_Kurabayashi1996, cooperative_map_making_Singh1993, Stroupe2004}. These works tackle problems concerning efficient exploration of an environment, like sweeping~\cite{cooperative_sweeping_Kurabayashi1996}, exploration with marsupial teams~\cite{coordinated_marsupial_teams_Burgard2010}, exploring using segmentation~\cite{coordinated_segmentation_Burgard2008}, etc. They all, in some way, deal with the mapping of an unknown environment. Other works, like~\cite{multi_robot_target_search_Dadgar2016, collaborative_target_search_Tang2021}, that deal with multi-robot collaborative target allocation also assume unknown environments. Our work differs in that we assume a partially known environment, where the only unknown is the location of the target.

\section{Problem formulation}\label{sec:problemFormulation}
We describe the discrete partially observable environment, the multi-agent system, and the control objectives.

\subsection{Environment model} \label{sec:env_model}
We consider a multi-agent system of $M$ agents with the aim of locating a target that is hidden in $\mathbb{R}^n$. Agents have some prior knowledge of where the target is located; they know that it is at one of a finite number of known \textit{goal nodes}, points in $\mathbb{R}^n$. To find the hidden target, agents can make observations in \textit{observation areas}, certain regions of state-space, and observations are shared between agents. Agents have a map of the environment, including goal nodes, observation areas, and potential obstacles. However, agents do not know at which of the goal nodes the target is located. As an example, see Figure~\ref{fig:experimentalSetup} where a quadruped and a drone are locating a science sample. In this example the goal nodes lie inside the blue observation areas, so agents can make observations near potential target locations. However, it is important to mention that observation areas can in general be placed anywhere, not only near goal nodes.

We assume a finite number, $G$, of goal nodes where the target can be located, corresponding to $G$ different environment states $e$. The environment state may change at each time step and it can only be inferred through noisy observations. The evolution of the environment state $e$, representing the goal location, is modeled using a \textit{hidden Markov model (HMM)}~\cite{krishnamurthy2016partially} given by the tuple $\mathcal{H} = \left( \mathcal{E}, \mathcal{O},T, Z \right)$:
\begin{itemize}

	\item $\mathcal{E}=\{0,1,\dots,G-1\}$ is a set of partially observable environment states.
    
	\item $\mathcal{O}=\{0,1, \dots,G-1\}$ is the set of observations for the partially observable state $e\in \mathcal{E}$.
			
    \item The function $T: \mathcal{E} \times \mathcal{E} \times \mathbb{R}^n \rightarrow [0,1]$ describes the probability of transitioning to a state $e'$ given the current environment state $e$ and system's state $x$, i.e.,
    $$
        T(e', e, x) :=\mathbb{P}(e'| e, x).
    $$

	\item The function $Z: \mathcal{E} \times \mathcal{O}\times \mathbb{R}^n \rightarrow [0,1]$ describes the probability of observing $o$, given the current environment state $e$ and the system's state $x$, i.e.,  
	$$
	    Z(e,o,x) :=  \Prob(o| e, x).
	$$
\end{itemize}

Finally, we define the belief vector $b_k \in \mathcal{B} = \{b \in \mathbb{R}^{|\mathcal{E}|}_{0+} : \sum_{i=0}^{|\mathcal{E}|-1}b[e]=1\}$, where each entry $b_k[e]$ represents the posterior probability, at time $k$, that the system's state $e_k$ equals to $e\in \mathcal{E}$.


\subsection{Control design objectives}
At time $k$ each agent $a$ tries to reach a way-point $x_{k+1}^a$, computed as
\begin{equation}\label{eq:sys}
    x^a_{k+1} = x^a_k + \Delta x^a_k,
\end{equation}
where $\Deltax^a_k \in \mathbb{R}^n$ is the way-point change that must satisfy $x^a_k + \Delta x^a_k \in \Xfree \eqdef \mathcal{X}^a \setminus \Xobs$, where $\mathcal{X}^a\in\mathbb{R}^n$ is the state-space and  $\Xobs\in\mathbb{R}^n$ the obstacle region, of agent $a$.

Let $j(k)$ be the number of observations made until time $k$ and $\mathbf{o}_{j(k)}=[o_1, \ldots, o_{j(k)}]$ the set of observations. As the environment is not perfectly known, and noisy observations are collected during execution, it is not ideal to compute a sequence of way-points offline. Instead, we compute a function $\pi^a_k$, which maps the environment observations up to time $k$ to the way-point change for agent $a$, i.e., 
\begin{equation} \label{eq:way_point_mapping}
    \Delta x^a_k = \pi^a_k(\mathbf{o}_{j(k)}).
\end{equation}

We denote the policy of a mission, ending at some time $N_p$, of agent $a$ by $\boldsymbol{\pi}^a=[\pi^a_0,\dots,\pi^a_{N_p-1}]$ and a plan $\boldsymbol{\pi}$ as the collection of all policies: $\boldsymbol{\pi}=\{\boldsymbol{\pi}^a\}_{a=1}^{M}$. A path is a realization of way-points $p^a = [x^a_0, x^a_1,\dots , x^a_{N_p}]$, such that $x^a_{k+1} = x^a_k + \Delta x^a_k, \forall k = 0,1,\dots,N_{p}-1$. $p=[p^1,\dots,p^M]$ is a collection of paths. Given a plan $\boldsymbol{\pi}$, we define the set of possible collections of paths under this plan by $\mathcal{P}_{\boldsymbol{\pi}}$. Note that, by \eqref{eq:way_point_mapping}, each $p\in\mathcal{P}_{\boldsymbol{\pi}}$ is a collection of paths corresponding to a realized set of observations agents make. 

Given the initial state $x(t)=[x^1(t),\dots,x^M(t)]$ and environment belief $b(t)$, 
as well as a plan $\boldsymbol{\pi}$, we define the cost of the plan for agent $a$, at time $t$, as follows:
\begin{equation} \label{eq:expected_cost}
\begin{aligned}
&J^a(x(t), b(t)) \\ &\qquad= \mathbb{E}_{\mathcal{P}_{\boldsymbol{\pi}}}\bigg[\sum_{k=0}^{N_p-1} h(x^a_k, \Deltax^a_k, e_k) + h_{N_p}(x^a_{N_p}, e_{N_p})\bigg| b(t)\bigg], 
\end{aligned}
\end{equation}
where the stage cost $h : \mathbb{R}^n \times \mathbb{R}^n \times \mathcal{E} \rightarrow \mathbb{R}$ and terminal cost $h_N : \mathbb{R}^n \times \mathcal{E} \rightarrow \mathbb{R}$ are functions of the partially observable environment state $e\in\mathcal{E}$, and the expectation is over the set of possible realizations of agent paths $\mathcal{P}_{\boldsymbol{\pi}}$. 



\section{The mixed observable RRT}
\label{sec:MORRT_section}

This section describes the main contribution of our work. We introduce the \textit{mixed observable RRT (MORRT)} and explain how it can be leveraged to compute a plan offline, given an environment map.

\subsection{Cost reformulation} \label{sec:cost_reform}
\begin{figure}[t]
    \centering
    \vspace{1mm}
    \includegraphics[width=0.75\linewidth, trim={0 0 0 0},clip]{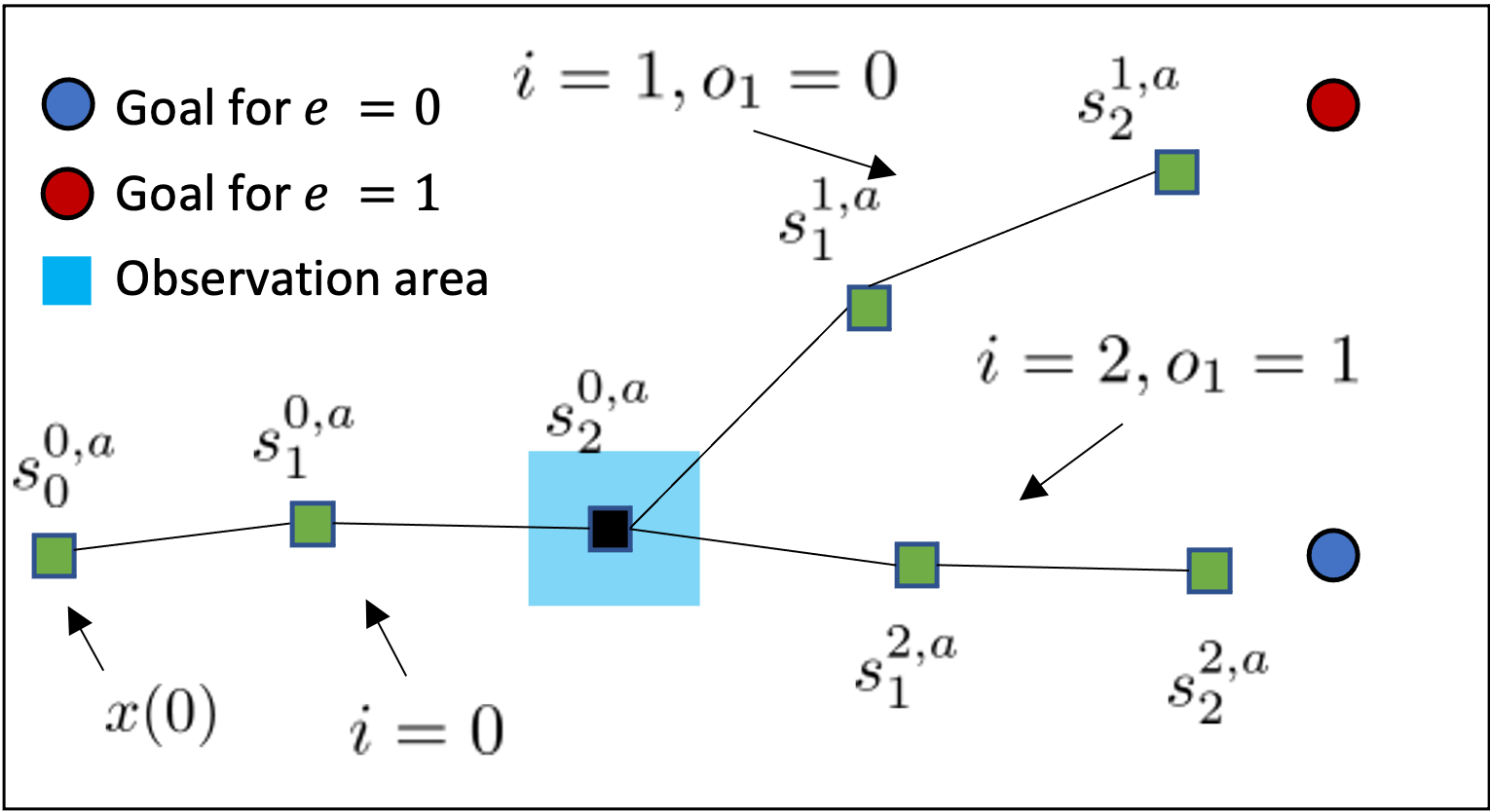} 
    \caption{A trajectory tree for agent $a$, representing a plan $\boldsymbol{\pi}$. Nodes and edges correspond to way-points and way-point changes, respectively, and each branch $i$ is associated with a unique observation vector. Here, $\mathcal{I}_{\boldsymbol{\pi}}=\{0,1,2\}$; $\Par(1)=\Par(2)=0$; $\Obs(0)=1$, and $\Obs(1)=\Obs(2)=0$. Branch 0 corresponds to the initial belief $b(0)$, and branches 1 and 2 correspond to the observation vectors $\mathbf{o}_1=[0]$ and $\mathbf{o}_1=[1]$, respectively.}
    \label{fig:branch_illustration_in_cost}
\end{figure}

The policy of agent $a$ can be illustrated by a trajectory tree where nodes and edges represent way-points and way-point changes, respectively. Each branch in the tree is a path that the agent commits to, given a specific observation and environment belief update. Basically, the controller plans different trajectories, and future decisions are a function of the realized environment observations. A plan, which is the collection of $M$ policies, can be illustrated as a collection of $M$ trajectory trees. The example in Figure~\ref{fig:branch_illustration_in_cost} consists of three trajectory branches; since observations are shared between agents, each agent in this example has a trajectory tree with three branches, corresponding to the branches in the figure. 

Denote the set of trajectory branches, for a given plan $\boldsymbol{\pi}$, by $\mathcal{I}_{\boldsymbol{\pi}}$. We introduce, for each branch $i\in\mathcal{I}_{\boldsymbol{\pi}}$ and each agent $a\in\{1,2,\dots,M\}$, the vector $s^{i,a}=[s^{i,a}_0,...,s^{i,a}_{N_i}]$, which given a plan $\boldsymbol{\pi}$, denotes the way-points of agent $a$ in the branch $i$, where $N_i+1$ is the number of way-points in this branch. It is important to note that, for each $s^{i,a}_\ell$, there is a one-to-one mapping to $x^a_k$ for some $k$, where $k$ is the time index; introducing the $s$-notation makes it easy to distinguish between different branches without keeping track of the time index $k$ in each branch. Moreover, let the function $\Par(i)$ return the parent branch of branch $i$, and let $\Obs(i)=1$ if branch $i$ ends at an \textit{observation node}---a way-point inside of an observation area---and $\Obs(i)=0$ otherwise (see Figure~\ref{fig:branch_illustration_in_cost} for an illustration).

Next, as the environment states are discrete the expected cost \eqref{eq:expected_cost} can be rewritten as a summation over way-points. We use Bayesian belief updates, i.e., after observing $o$ at time $k$ the belief is updated as
$$
b_k[e] = \frac{Z(e,o,x_k)}{\mathbb{P}(o|x_k,b_{k-1}[e])}\sum_{e'\in\mathcal{E}} T(e,e',x_k)b_{k-1}[e']
$$
for the functions $Z$ and $T$ defined in Section~\ref{sec:env_model}.

\begin{proposition}
Denote the state of the multi-agent system at time $k$ as $x_k=[x^1_k,...,x^M_k]$. Let $o_i$ denote the observation corresponding to the branch $i\in\mathcal{I}_{\boldsymbol{\pi}}$, and $v^{i}[e] =\Theta(o_i, s_0^{i})v^{\Par(i)}[e]$ the unnormalized belief, where
\begin{equation*}
\Theta(o_i, s_0^{i}) = \text{diag}\big(\begin{matrix} Z(0, o_i, s_0^{i}) & \ldots & Z(|\mathcal{E}|-1, o_i, s_0^{i})
\end{matrix}\big).
\end{equation*} 
The expected cost~\eqref{eq:expected_cost} can be expressed equivalently as 
\begin{equation} \label{eq:cost_as_sum}
\begin{aligned}
&J^a(\mathcal{I}_{\boldsymbol{\pi}}, x(t), b(t)) 
=
\sum_{i\in \mathcal{I}_{\boldsymbol{\pi}}} \Bigg\{ \sum_{\ell=0}^{N_i-1}   \sum_{e \in \mathcal{E}}  v^{i}[e] h(s_\ell^{i, a}, \Deltas_\ell^{i, a}, e) 
\\ 
& \qquad\qquad\qquad 
+  \mathds{1}[\Obs(i)=0] \sum_{e \in \mathcal{E}}  v^{i}[e]  h_{N}(s_{N_i}^{i, a}, e)\Bigg\}, 
\end{aligned}
\end{equation}
where $s^0_0=x(t)$.
\end{proposition}
\proof
This is a direct extension of Proposition~1 in~\cite{rosolia2021mixedobservable}: instead of making observations at each time step $k=1,2,\dots$, agents make observations at the beginning of each branch $i\in\mathcal{I}\setminus \{0\}$, which amounts to a summation over the nodes in each branch, rather than each time step.
\endproof

\begin{figure*}[t]
    \centering
    \subfloat[Root RRT.]{\includegraphics[width=0.18\linewidth]{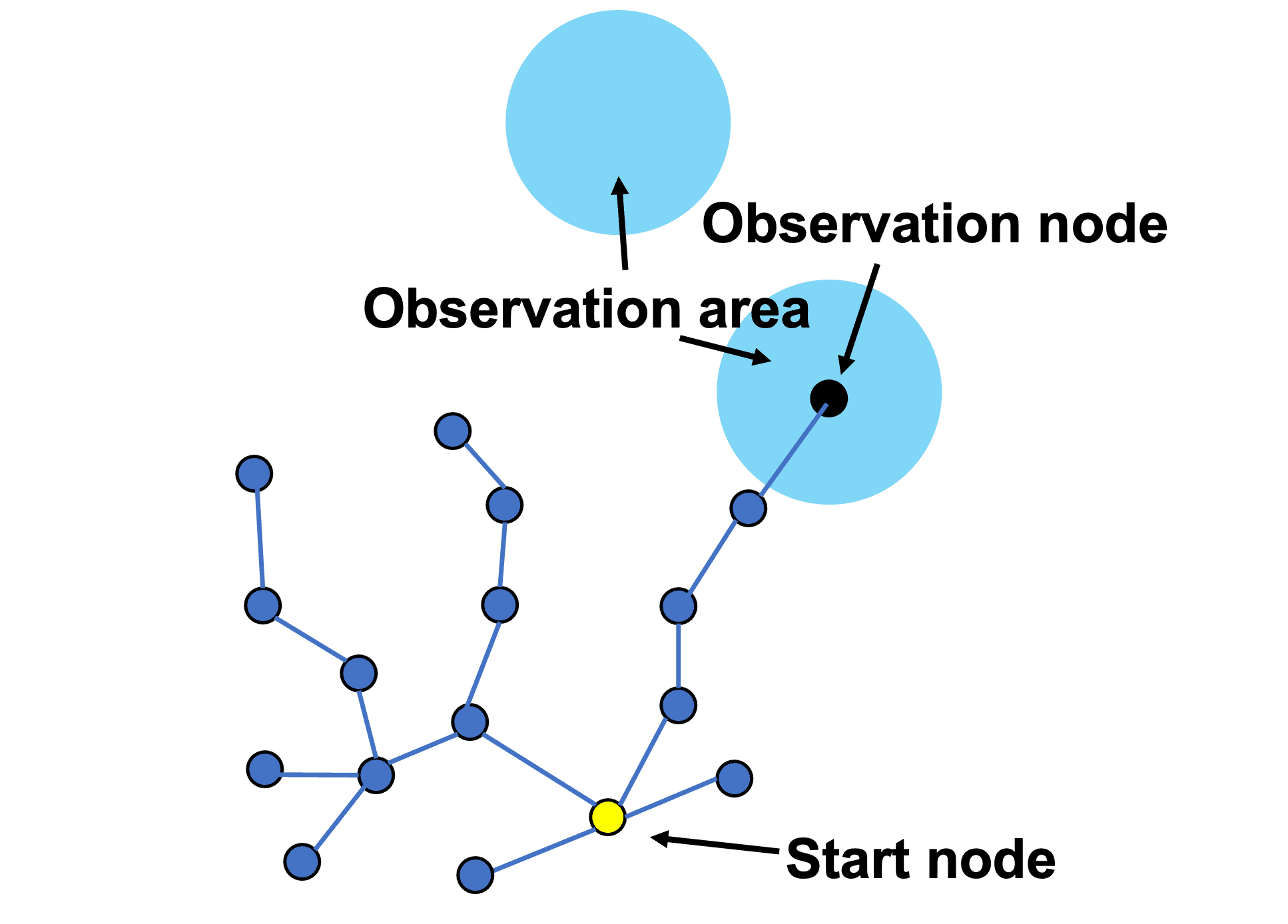}\label{subfig:MORRT_a}} \hspace{3.5mm}
   \subfloat[RRT from first observation area.]{ \includegraphics[width=0.18\linewidth]{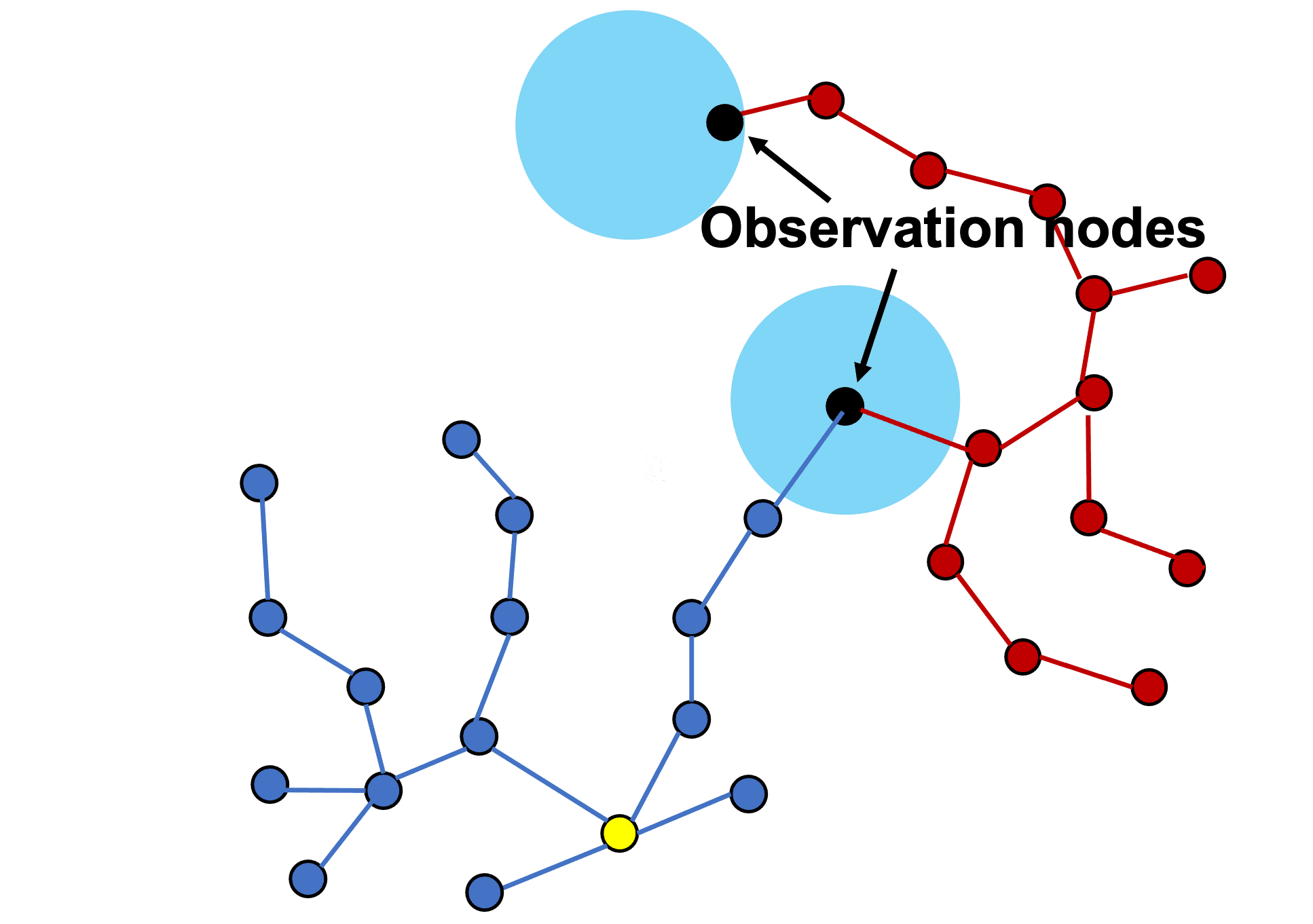}\label{subfig:MORRT_b}} \hspace{3.5mm}
    \subfloat[RRT from second observation area.]{\includegraphics[width=0.18\linewidth]{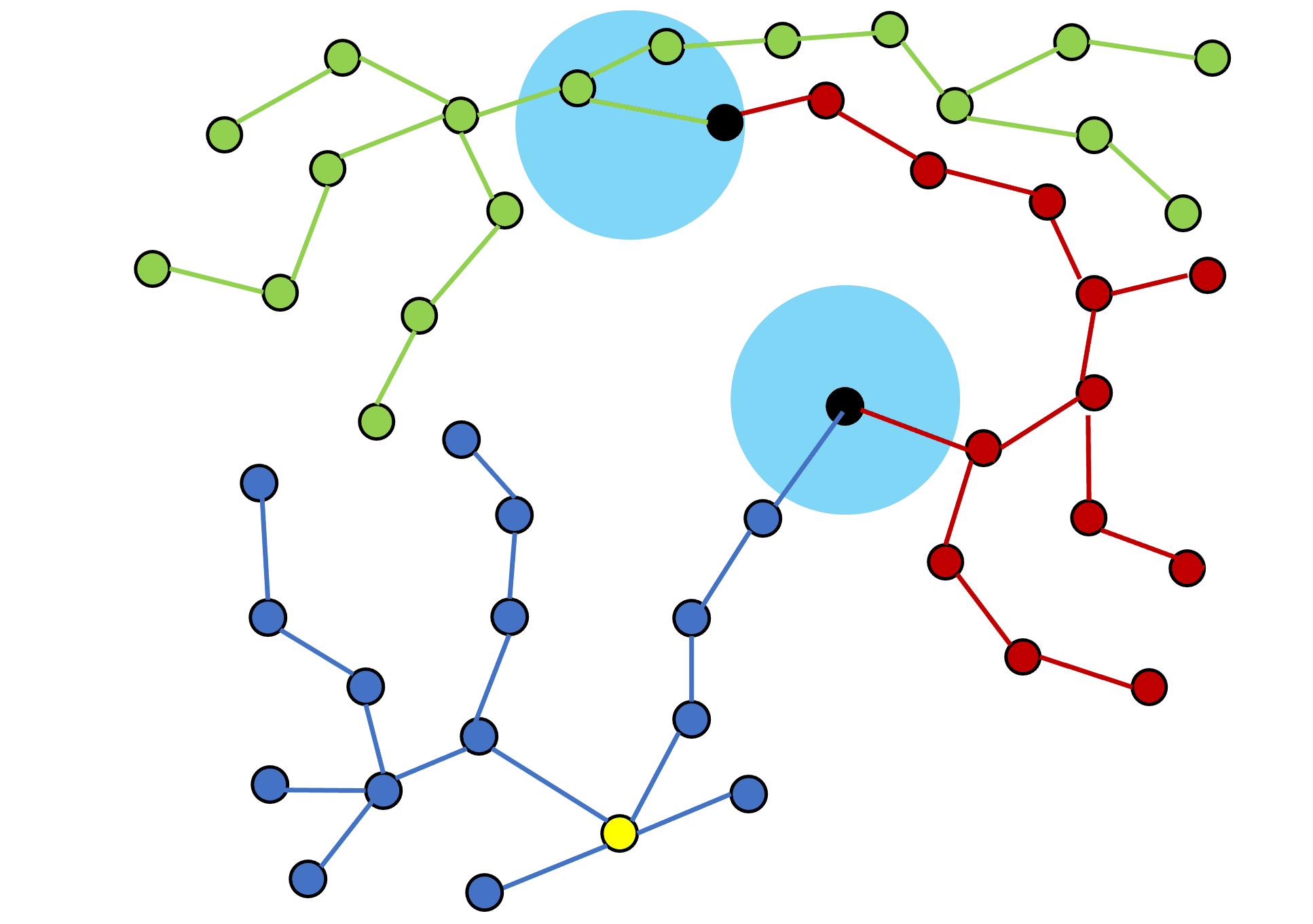}\label{subfig:MORRT_c}} \hspace{3.5mm}
    \subfloat[Example path.]{\includegraphics[width=0.18\linewidth]{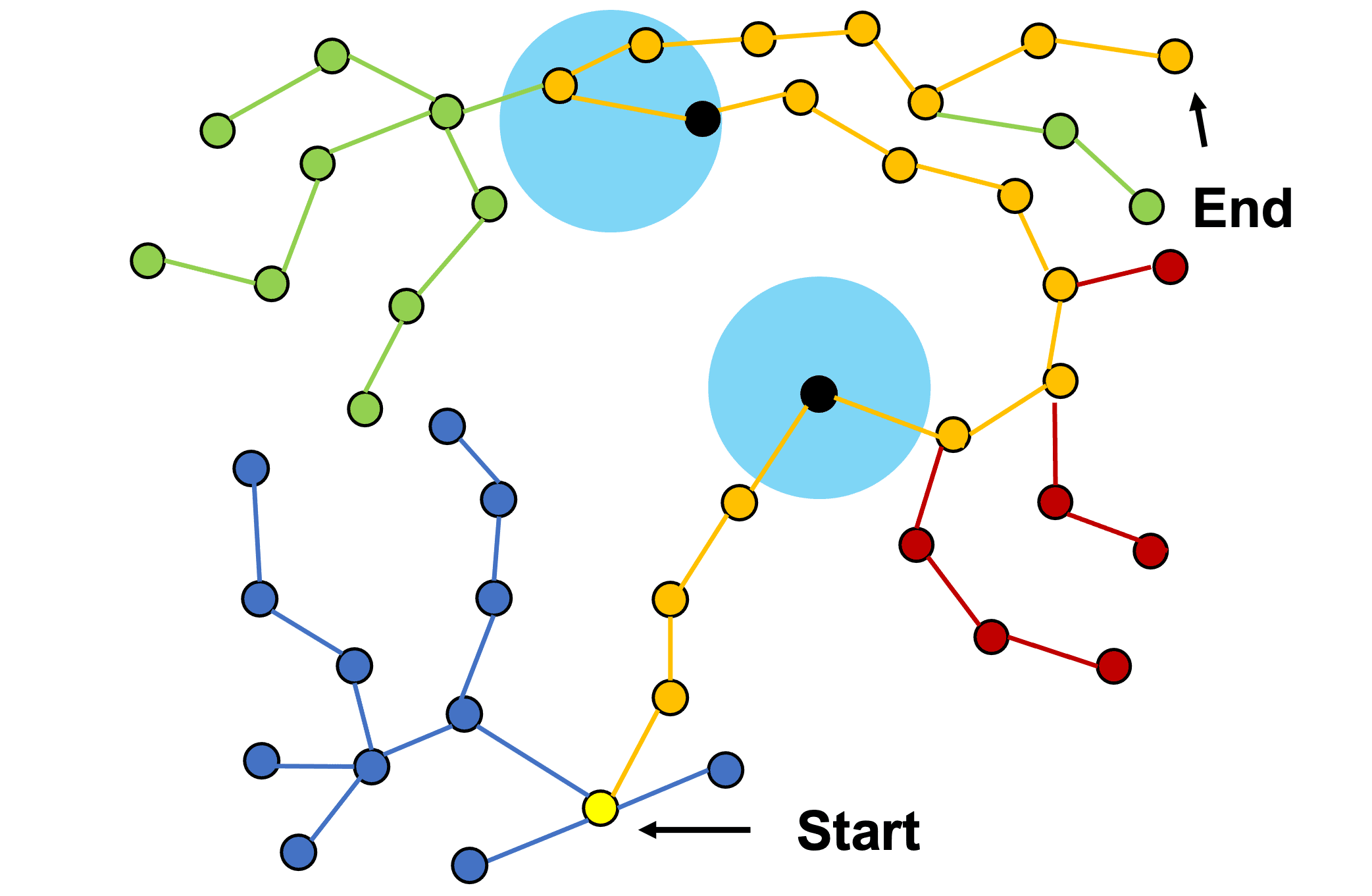}\label{subfig:MORRT_d}}
    \caption{The MORRT expansion for $\Nobs=1$. The MORRT expands from left to right. Blue, red, and green nodes belong to three separate RRTs. The yellow path to the right shows one possible path in the MORRT.}
    \label{fig:build_plan_illustration}
\end{figure*}

\subsection{Building the MORRT} \label{sec:buildTree}
\begin{algorithm}[b]
\footnotesize
\caption{\footnotesize RRT$\leftarrow\texttt{RRT}(\mathcal{T}, \Deltax, N, K$)}
\label{alg:RRT}
    \begin{algorithmic}[1]
 \REPEAT
 \STATE $x_\textrm{rand}\leftarrow \texttt{random\_state}()$; $x_{\textrm{near}}\leftarrow \texttt{nearest}(x_{\textrm{rand}},\mathcal{T})$
    \STATE $x_\textrm{new}\leftarrow \texttt{new\_node}(x_\textrm{near},x_\textrm{rand},\Deltax)$; $\mathcal{T}.\texttt{add}(x_\textrm{new})$ 
    \STATE $x_{\textrm{new}}.\texttt{count}\leftarrow x_{\textrm{near}}.\texttt{count}+1$; $x_{\textrm{new}}.\texttt{parent}\leftarrow x_{\textrm{near}}$
    \IF{$x_{\textrm{new}}$ \textbf{in} an area $\textbf{in } \mathcal{T}.\texttt{areas}$}
    \STATE $\mathcal{T}.\texttt{x\_obs}.\texttt{append}(x_\textrm{new})$; $x_{\textrm{new}}.\texttt{area}\leftarrow\texttt{area}$
    \ENDIF
\UNTIL{$N$ observations or $K$ nodes}
    \end{algorithmic}
\end{algorithm}

We now show how to construct an MORRT for the single-agent case. The multi-agent extension is discussed in Section~\ref{sec:multi_agent_extension}. First, we introduce a modified RRT that stores observations collected in observation areas. We leverage this RRT algorithm to construct the MORRT. Algorithm~\ref{alg:RRT} constructs the RRT while storing observations. It takes as input a tree $\mathcal{T}$ and three parameters $\Deltax,N,$ and $K$. Each tree $\mathcal{T}$ has three features: $\mathcal{T}.\texttt{x\_obs}$, a set of all observation nodes in $\mathcal{T}$; $\mathcal{T}.\texttt{areas}$, a set of observation areas; and $\mathcal{T}.\texttt{child}$, the child RRTs in a tree of RRTs. Moreover, each node $x\in\mathcal{T}$ has three features: $x.\texttt{area}$, an observation area, if $x$ is inside of one; $x.\texttt{count}$, the path length from the root node (henceforth denoted $x_{\textrm{init}}$) to $x$ in $\mathcal{T}$; and $x.\texttt{parent}$, the parenting node in $\mathcal{T}$. In line 2 of Algorithm~\ref{alg:RRT}, a node $x_{\textrm{rand}}$ is sampled in state-space and the nearest node to $x_{\textrm{rand}}$ in $\mathcal{T}$ is denoted $x_{\textrm{near}}$. The $\texttt{new\_node}$ function (line 3) generates a new node  $x_{\textrm{new}}$ between $x_{\textrm{rand}}$ and $x_{\textrm{near}}$, a distance $\Deltax$ from $x_{\textrm{near}}$ (considering obstacles); $\texttt{add}$ adds this node to the tree. In line 4 the \texttt{count} and \texttt{parent} of $x_{\textrm{new}}$ are defined. If $x_{\textrm{new}}$ is inside an observation area that is in $\mathcal{T}.\texttt{areas}$, the node is added to the set of observation nodes and the area is stored as $x_{\textrm{new}}.\texttt{area}$ (lines 5--6). Lines 1--7 repeat until the tree $\mathcal{T}$ has $N$ observation nodes---nodes in observation areas---or $K$ total nodes, whichever happens first.\footnote{RRT$^{*}$~\cite{KaramanFrazzoliRRTstar} can be leveraged by introducing a \textit{cost heuristic} for a path, like the cost of committing to this path regardless of observations.}

The MORRT is illustrated in Algorithm~\ref{alg:MORRT}, which constructs a tree of RRTs where the nodes are observation nodes inside of observation areas, and the edges are standard RRTs that extend between these observation nodes. In line 1 the \texttt{count} of $x_\textrm{init}$ is set to zero. In line 2 a ``root RRT'' is initialized at $x_{\textrm{init}}$, with the set of observation nodes initially empty, and the set of (available observation) areas initialized as $\texttt{areas}$. In line 3 the root RRT is extended with Algorithm~\ref{alg:RRT}. Once the root RRT has been expanded it is placed in a set of \texttt{parents} (line 4). From each observation node in the parent RRT (line 6), a ``child RRT'' is initialized with an empty set of observation nodes (line 7). The child RRT's set of available observation areas is the parent's \texttt{areas} minus the observation area of $x_{\textrm{obs}}$ (line~8); with this construction, the further up in the tree of RRTs we get, the fewer observation areas become available to explore, until eventually no more areas will be explored following line 5 in Algorithm~\ref{alg:RRT}. The child RRT is expanded with the \texttt{RRT} function (line 9). Finally, the child RRT is appended to the set of parents (and considered as a parent in a following iteration) and defined as a child of the parent (line 10). Lines 5--10 repeat until there is no parent RRT with observation nodes, i.e, until $\mathcal{T}_{\textrm{parent}}.\texttt{x\_obs}$ in line 6 is empty for all remaining parents. Note that the algorithm will terminate since the set $\mathcal{T}$.\texttt{areas} shrinks as the depth in the tree of RRTs increases, and there is only a finite number of observation areas. Eventually $\mathcal{T}.\texttt{x\_obs}$ will be empty for all $\mathcal{T}$ deep enough in the tree of RRTs, following lines 7--8 of Algorithm~\ref{alg:MORRT} and lines 5--6 of Algorithm~\ref{alg:RRT}.

Figure~\ref{fig:build_plan_illustration} illustrates Algorithm~\ref{alg:MORRT} with $N=1$. An RRT is expanded until it reaches an observation area (Figure~\ref{subfig:MORRT_a}). Since $N=1$ a new RRT is immediately initialized from this observation node and expanded until it reaches the second observation area (Figure~\ref{subfig:MORRT_b}). Once all observation areas have been visited, a final RRT of fixed size is expanded from the most recently visited observation area (Figure~\ref{subfig:MORRT_c}), which follows from the fact that line 5 of Algorithm~\ref{alg:RRT} will never hold for this RRT, so the tree will expand to $K$ nodes. Figure~\ref{subfig:MORRT_d} shows a possible path through the MORRT; however, all paths ending at a node without children are potential paths.

\begin{algorithm}[t!]
\footnotesize
\caption{\footnotesize $\mathcal{T}_\textrm{root}\leftarrow\texttt{MORRT}(x_\textrm{init}, \Deltax, \Nobs, K$, $\texttt{areas})$}
\label{alg:MORRT}
\begin{algorithmic}[1]
\STATE $x_{\textrm{init}}.\texttt{count}\leftarrow 0$
\STATE $\mathcal{T}_{\textrm{root}}.\texttt{init}(x_{\textrm{init}})$;  $\mathcal{T}_{\textrm{root}}.\texttt{x\_obs}\leftarrow \{\}$; $\mathcal{T}_{\textrm{root}}.\texttt{areas}\leftarrow \texttt{areas}$
\STATE $\mathcal{T}_{\textrm{root}}$ $\leftarrow$ \texttt{RRT($\mathcal{T}_{\textrm{root}},\Deltax,N,K$)} \texttt{//ALGO~\ref{alg:RRT}}
\STATE \texttt{parents} = $\{\mathcal{T}_{\textrm{root}}\}$
\FOR{\textbf{each} $\mathcal{T}_{\textrm{parent}}$ \textbf{in} \texttt{parents}}
\FOR{\textbf{each} $x_{\textrm{obs}}$ \textbf{in} $\mathcal{T}_{\textrm{parent}}.\texttt{x\_obs}$}
\STATE $\mathcal{T}_{\textrm{child}}.\texttt{init}(x_{\textrm{obs}})$;  $\mathcal{T}_{\textrm{root}}.\texttt{x\_obs}\leftarrow \{\}$   
\STATE $\mathcal{T}_{\textrm{child}}.\texttt{areas}\leftarrow \mathcal{T}_{\textrm{parent}}.\texttt{areas}\setminus \{x_{\textrm{obs}}.\texttt{area}\}$

\STATE $\mathcal{T}_{\textrm{child}} \leftarrow$ \texttt{RRT($\mathcal{T}_{\textrm{child}},\Deltax,N,K$)} \texttt{//ALGO~\ref{alg:RRT}}
\STATE parents.\texttt{append($\mathcal{T}_{\textrm{child}}$)}; $\mathcal{T}_{\textrm{parent}}.\textrm{child}\leftarrow \mathcal{T}_{\textrm{child}}$
\ENDFOR
\ENDFOR

\end{algorithmic}
\end{algorithm}



\textbf{Selecting the best plan from the MORRT.}
The MORRT (Algorithm~\ref{alg:MORRT}) generates a tree of RRTs as shown in Figure~\ref{fig:build_plan_illustration}. A mission plan is a sub-tree of RRTs, where each RRT originates from an observation node and has at most $|\mathcal{O}|$ branches, as shown in Figure~\ref{fig:branch_illustration_in_cost}. The objective now is to find the sub-tree of RRTs that minimizes the cost~\eqref{eq:cost_as_sum}, which is done with dynamic programming on the tree of RRTs. We start from the latest trees in the tree of RRTs and evaluate the cost \eqref{eq:cost_as_sum}. Then, moving backwards we compute the cost in a dynamic programming manner through the tree of RRTs, while storing the best plan from each observation node.

\subsection{Multi-agent extension} \label{sec:multi_agent_extension}
\begin{algorithm}[b]
\footnotesize
\caption{\footnotesize MA-RRT$\leftarrow\texttt{MA-RRT}(M,x_\textrm{init}, \Deltax, N, K$)}
\label{alg:generate_multiAgent_RRT}
    \begin{algorithmic}[1]
    \FOR{$a=1,2,\dots,M$}
    \STATE $\mathcal{T}_a \leftarrow \texttt{RRT}(x_{\textrm{init}}[a],\Deltax,N,K)$ \texttt{//ALGO~\ref{alg:RRT}}
\ENDFOR
\STATE MA-RRT $\leftarrow \{\mathcal{T}_1,...,\mathcal{T}_M\}$

\STATE $\texttt{observations}\leftarrow \mathcal{T}_1.\texttt{x\_obs}\times
\mathcal{T}_2.\texttt{x\_obs}\times \dots \times \mathcal{T}_M.\texttt{x\_obs}$ 

\FOR{\textbf{each} \texttt{x\_obs} \textbf{in} \texttt{observations}}
\STATE $L\leftarrow \min\{x.\texttt{count} \textrm{ for } x\in\texttt{x\_obs}\}$
\FOR{\textbf{each} $x$ \textbf{in} \texttt{x\_obs}}
\WHILE{$x.\texttt{count} > L$}
\STATE $x \leftarrow x.\texttt{parent}$
\ENDWHILE
\ENDFOR
\ENDFOR
\STATE MA-RRT$.\texttt{x\_obs}\leftarrow \texttt{observations}$
    \end{algorithmic}
\end{algorithm}

The only modification in the multi-agent case is how the RRT is expanded. In the multi-agent extension, the $\texttt{RRT}$ function, lines 3 and 9 of Algorithm~\ref{alg:MORRT}, is substituted with Algorithm~\ref{alg:generate_multiAgent_RRT}---a multi-agent extension of the RRT. In lines 1--2 of Algorithm~\ref{alg:generate_multiAgent_RRT} $M$ separate RRTs are generated with the \texttt{RRT} function (Algorithm~\ref{alg:RRT}). The collection of the $M$ agent's RRTs defines the \textit{multi-agent RRT (MA-RRT)} (line~3). All combinations of the agent's observation nodes are stored in an \texttt{observations} set (line 4). This set is modified in lines 5--9 to make all agent's paths to the observation the same length. Each combination \texttt{x\_obs} of observation nodes (line 5) consists of $M$ single-agent observation nodes. We define $L$ as the minimum \texttt{count} variable of these nodes (line 6) (recall that, for an agent $a$, $x.\texttt{count}$ denotes the distance from $x_{\textrm{init}}[a]$ to $x$ in $\mathcal{T}_a$). Line 7 loops through all observation nodes $x\in\texttt{x\_obs}$, and lines 8--9 replaces $x$ with its parenting node until $x.\texttt{count}$ equals $L$. The MA-RRT observation nodes is the set of modified observations (line 10). Now, all agents have the same path length to the observations.

\begin{figure}[b]
    \centering
    \includegraphics[width=0.75\linewidth]{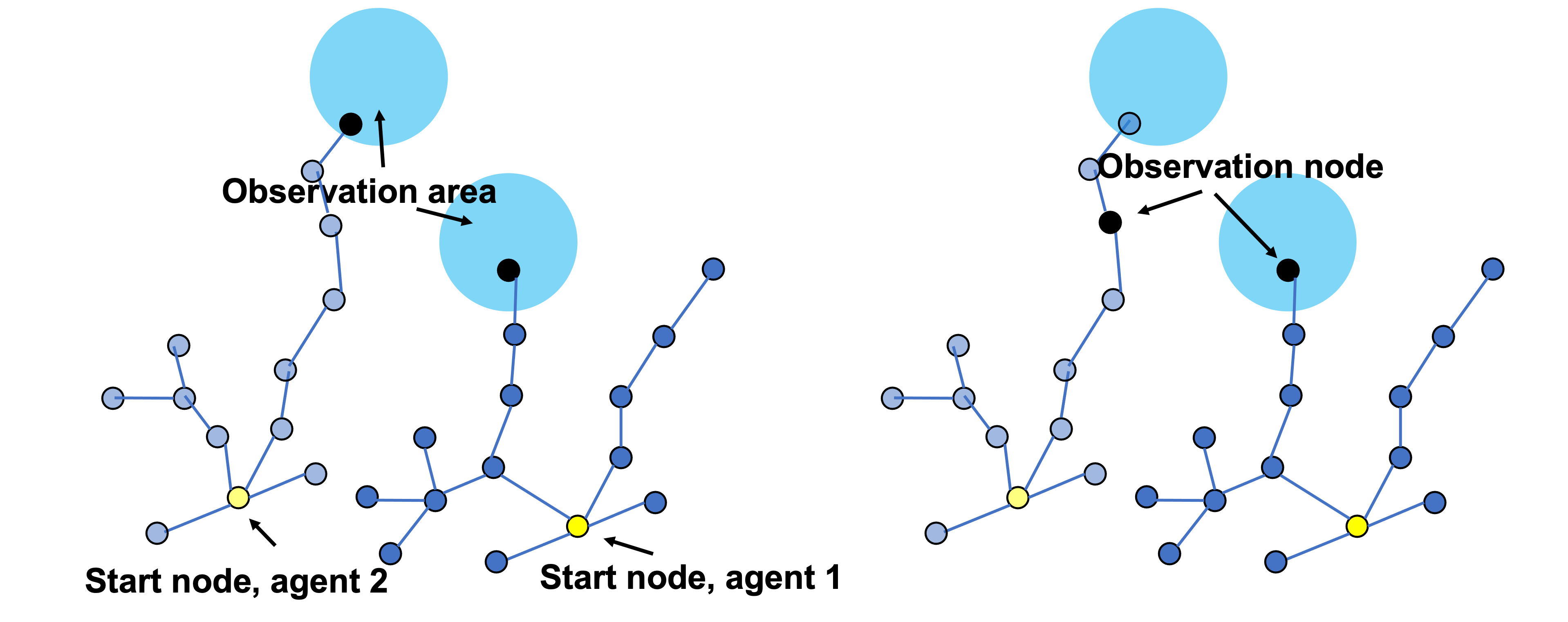} 
    \caption{Illustration of lines 5--9 of Algorithm~\ref{alg:generate_multiAgent_RRT}}
    \label{fig:MultiAgentMorrt_illustration}
\end{figure}

Figure~\ref{fig:MultiAgentMorrt_illustration} illustrates lines 5--10 of Algorithm~\ref{alg:generate_multiAgent_RRT} for $M=2$ (two agents) and $\Nobs=1$. The observation node of agent 2 is moved two nodes upwards in the path, to make both agent's paths from start to the observation node the same length.

\textbf{Complexity.} The MORRT complexity can be expressed in terms of the number of RRTs that are built. Let $\Noa$ denote the number of observation areas. The MORRT builds a tree of RRTs, and the number of branches originating from each node is at most $\Nobs^M$, where $\Nobs$ is a parameter of Algorithm~\ref{alg:MORRT}. Hence, the number of RRTs needed for an MORRT is at most $(\Nobs^M)^{\Noa} = \Nobs^{M\times  \Noa}$, which is a function of the number of agents $M$, the number of observation areas $\Noa$, and the parameter $\Nobs$. Computational time can be reduced by, for each agent, only keeping one observation node in each observation area before taking the Cartesian product in line 4 of Algorithm~\ref{alg:generate_multiAgent_RRT}; the complexity would then be $\Noa^{M\times  \Noa}$, which reduces computational time when $\Noa<\Nobs$. To reduce complexity in large systems we can assign different observation areas to different agents. This can be done randomly over several rounds, before selecting the best assignment, or deterministically if a good assignment is known beforehand. We illustrate this in Section~\ref{sec:five_agent_example}.

\begin{figure}[t]
    \centering
    {\includegraphics[width=0.9\linewidth]{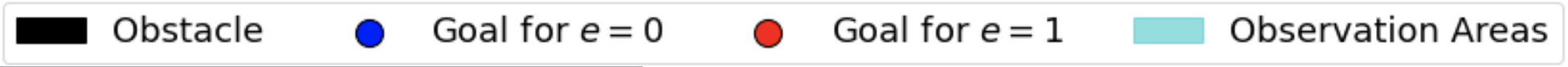}} \\ \vspace{-10pt}
    \subfloat[$\mathbf{o}_{3}=(0,0,0)$]{\includegraphics[width=0.33\linewidth]{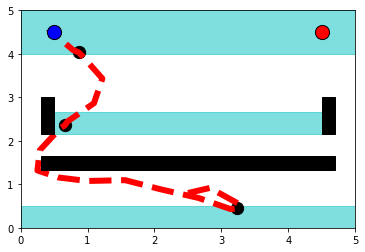}\label{subfig:singleAgent_path1}}
    \subfloat[$\mathbf{o}_{3}=(0,0,1)$]{\includegraphics[width=0.33\linewidth]{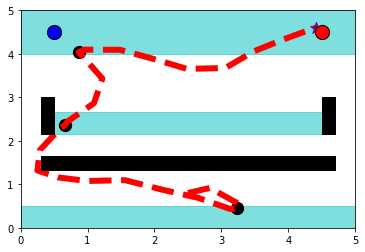}\label{subfig:singleAgent_path2}}
    \subfloat[$\mathbf{o}_{3}=(0,1,0)$]{\includegraphics[width=0.33\linewidth]{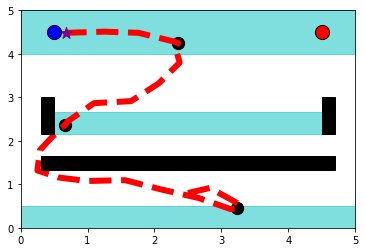}\label{subfig:singleAgent_path3}} \\
    \subfloat[$\mathbf{o}_{3}=(0,1,1)$]{\includegraphics[width=0.33\linewidth]{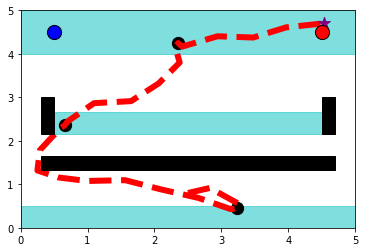}\label{subfig:singleAgent_path4}}
    \subfloat[$\mathbf{o}_{3}=(1,0,0)$]{\includegraphics[width=0.33\linewidth]{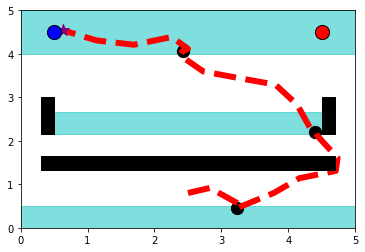}\label{subfig:singleAgent_path5}}
    \subfloat[$\mathbf{o}_{3}=(1,1,0)$]{\includegraphics[width=0.33\linewidth]{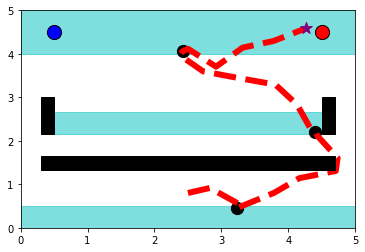}} \\
    \subfloat[$\mathbf{o}_{3}=(1,0,1)$]{\includegraphics[width=0.33\linewidth]{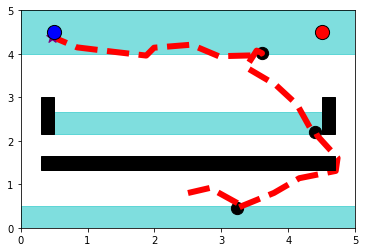}}
    \subfloat[$\mathbf{o}_{3}=(1,1,1)$]{\includegraphics[width=0.33\linewidth]{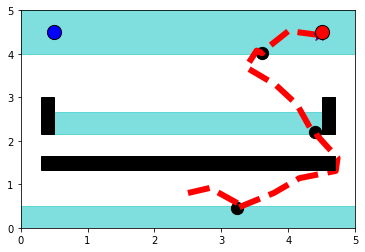}\label{subfig:singleAgent_path8}}
     \subfloat[Benchmark plan.]{\includegraphics[width=0.33\linewidth, trim={35 20 35 35},clip]{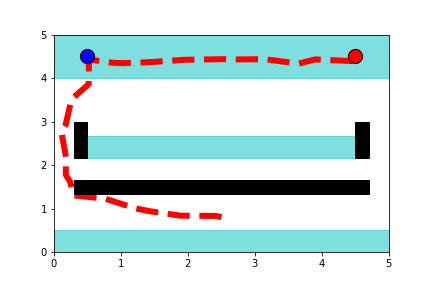}\label{subfig:benchmark}}
    
    \caption{Plan for an agent initiated below the obstacle. Black dots represent observation nodes.}
    \label{fig:singleAgentExample}
\end{figure}

\begin{figure}[!b]
    \centering
    \subfloat[\scriptsize{$\mathbf{o}_2=(0,0)$.}]{\includegraphics[width=0.67\linewidth]{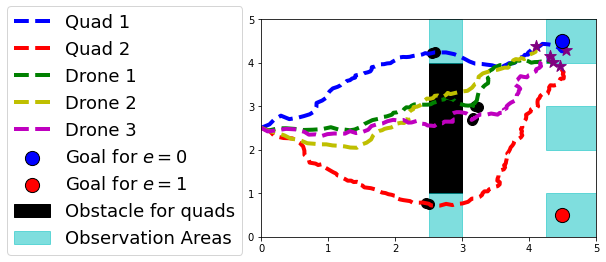}\label{subfig:multi_agent_1}} \\ 
    \subfloat[\scriptsize{$\mathbf{o}_4=(0,1,0,0)$.}]{\includegraphics[width=0.37\linewidth]{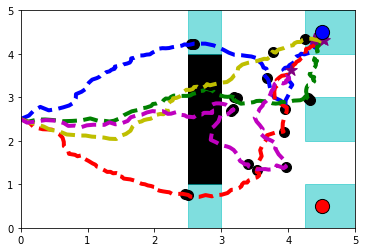}\label{subfig:multi_agent_2}} 
    \subfloat[\scriptsize{$\mathbf{o}_5=(0,1,0,1,0)$.}]{\includegraphics[width=0.37\linewidth]{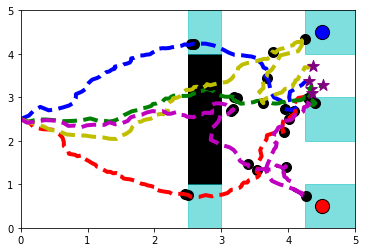}\label{subfig:multi_agent_3}} 
    \caption{Mission plan for two quadrupeds and three drones. Purple stars illustrate the end of a mission, and black dots represent observation nodes.}
    \label{fig:multi_agent_simulation}
\end{figure}

\begin{figure*}[t!]
    \centering
    \includegraphics[width=0.19\linewidth]{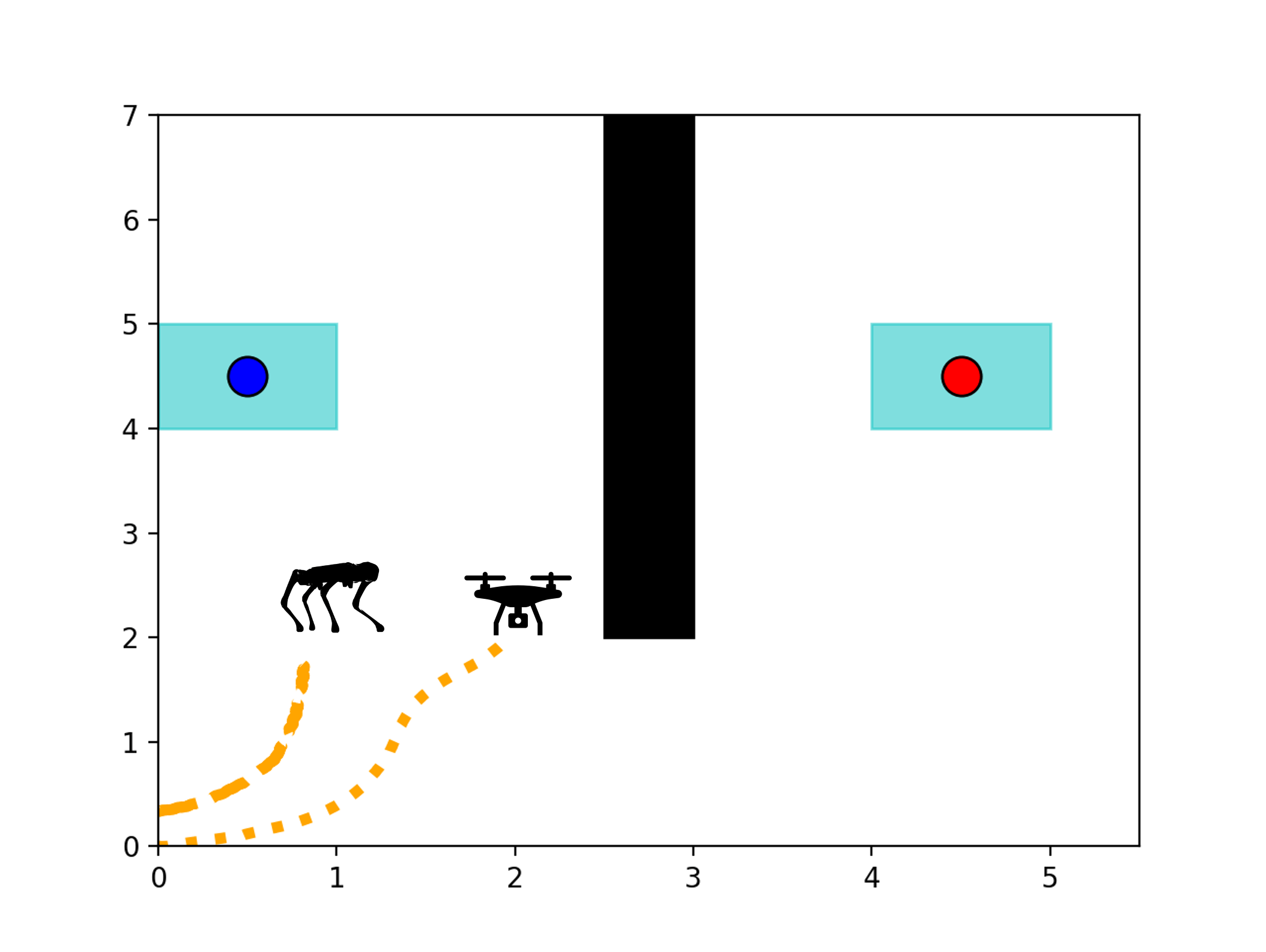}
    \includegraphics[width=0.19\linewidth]{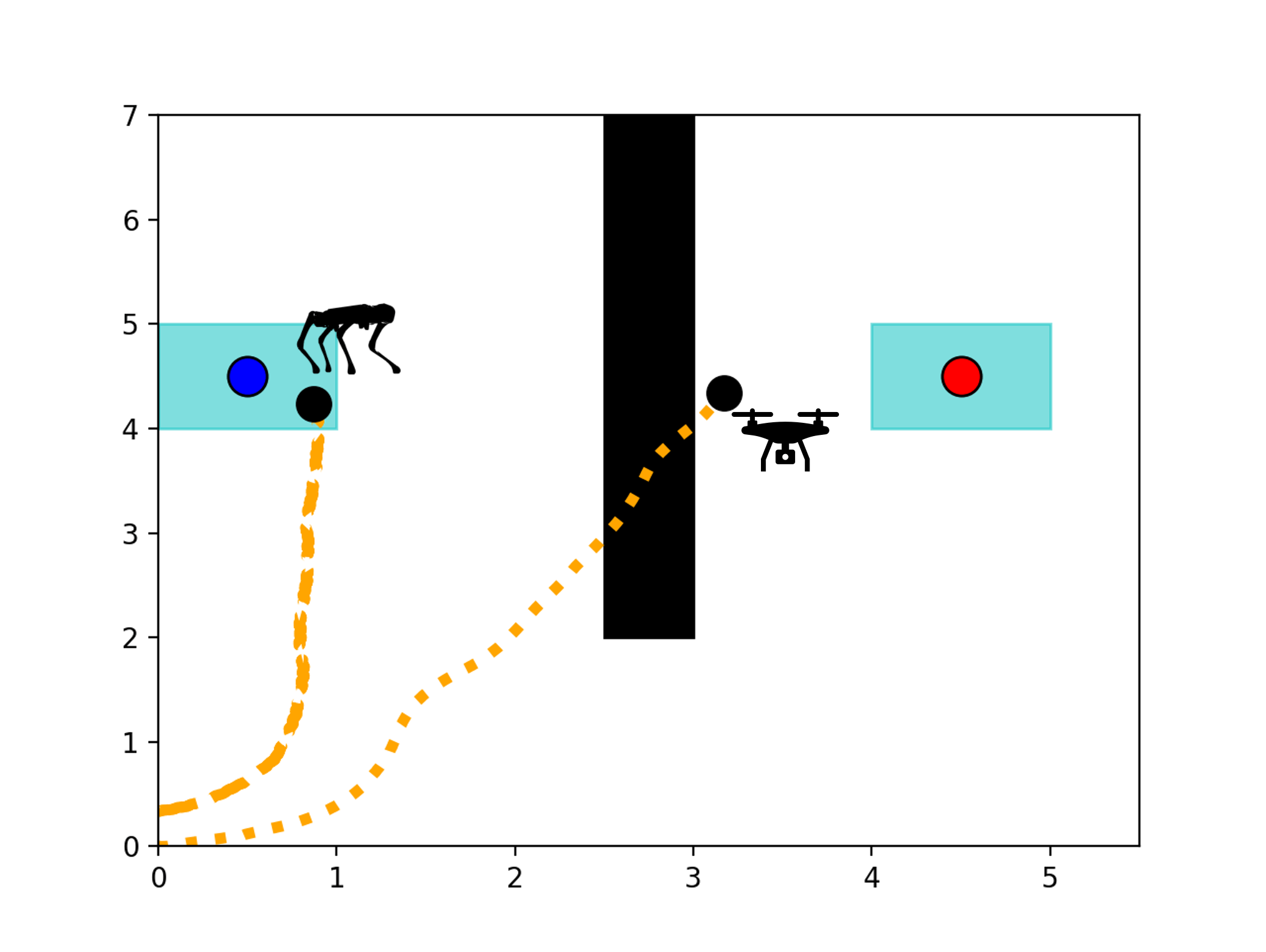}
    \includegraphics[width=0.19\linewidth]{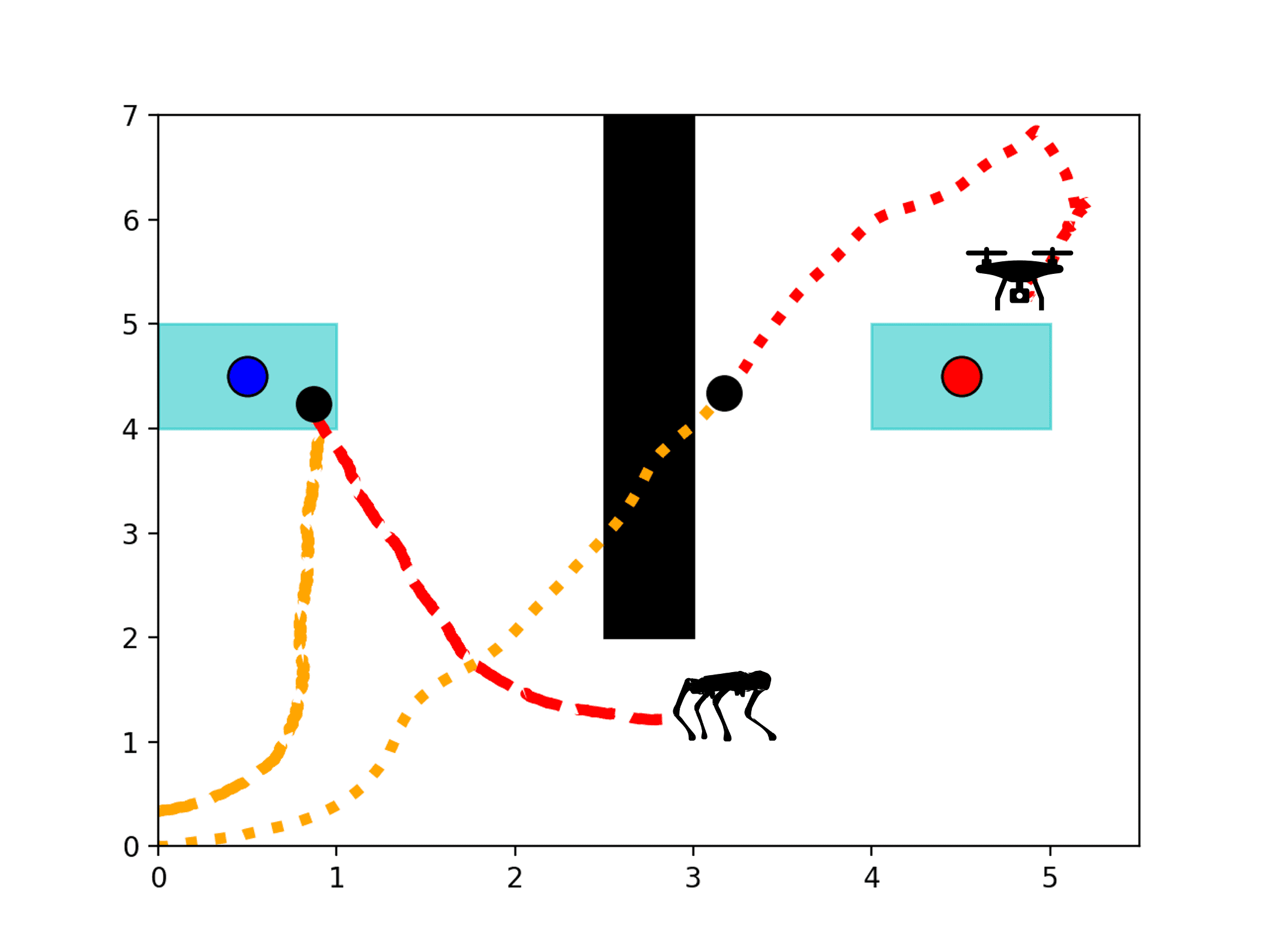}
    \includegraphics[width=0.19\linewidth]{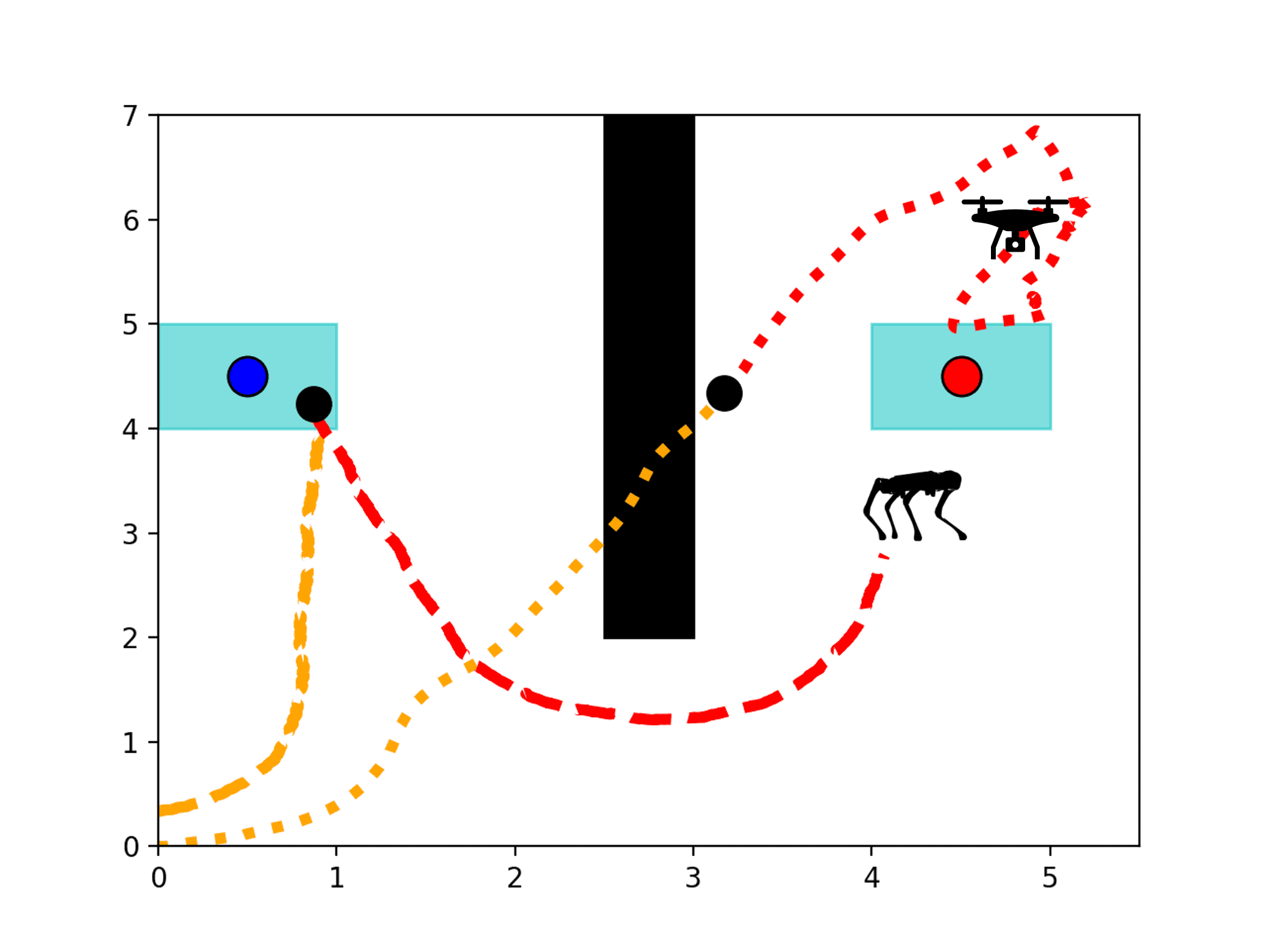}
    \includegraphics[width=0.19\linewidth]{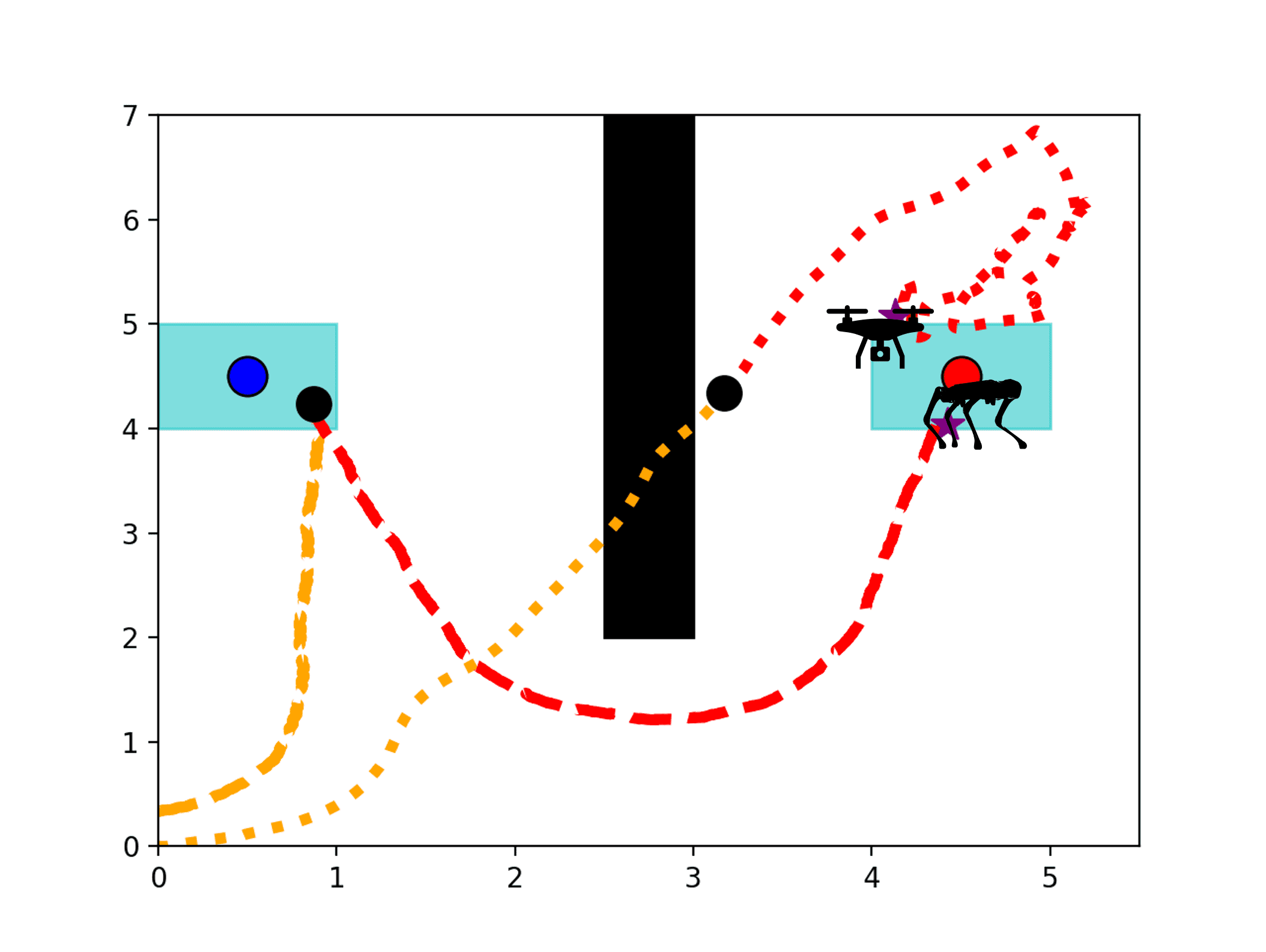} \\

    \includegraphics[width=0.19\linewidth, trim={25 10 25 25},clip]{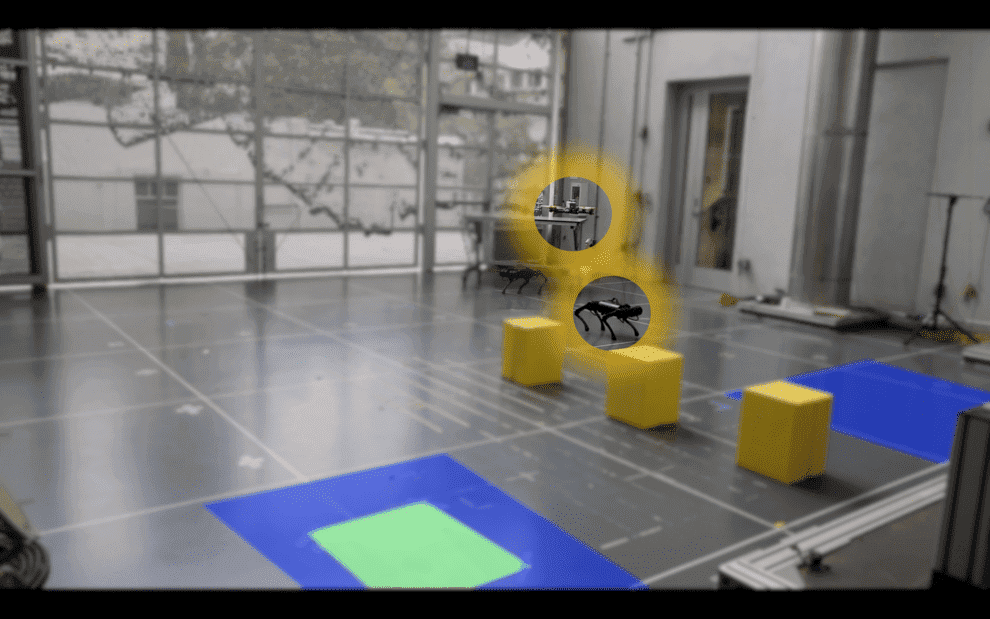} 
    \includegraphics[width=0.19\linewidth, trim={15 10 10 10},clip]{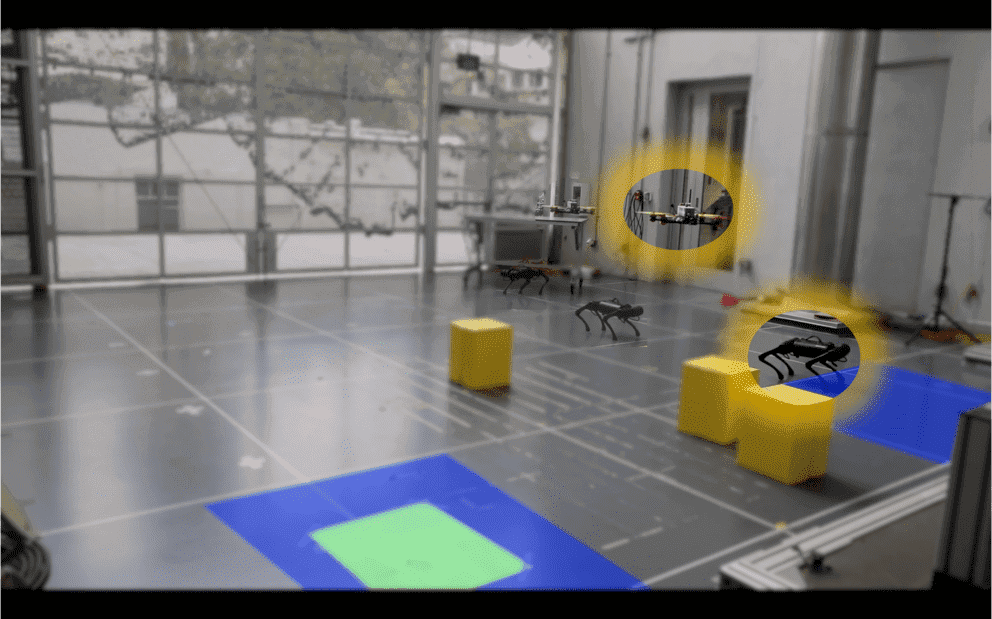} 
    \includegraphics[width=0.19\linewidth, trim={9 10 10 10},clip]{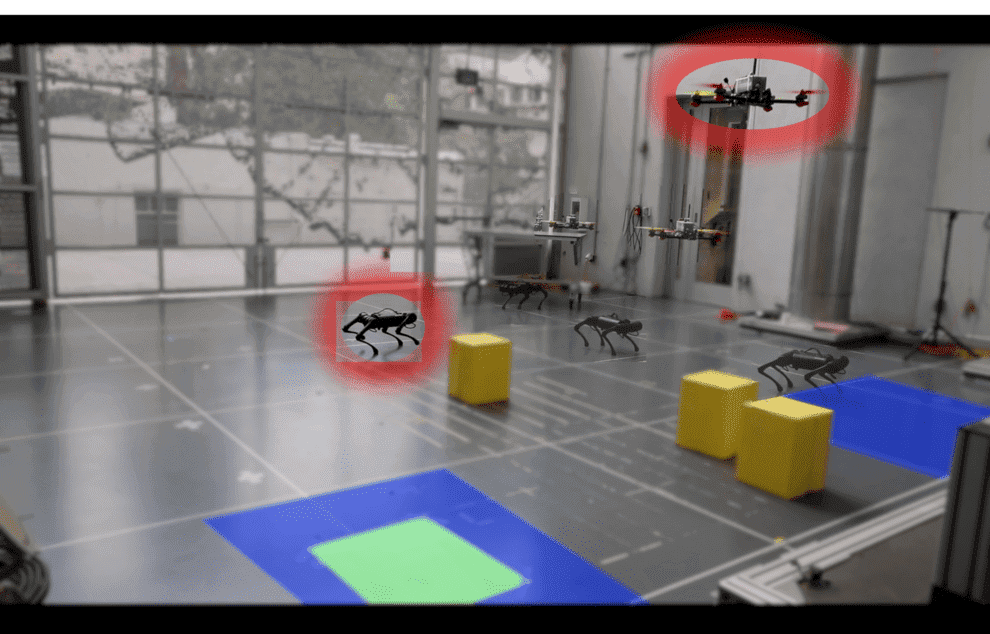} 
    \includegraphics[width=0.19\linewidth, trim={20 25 25 23},clip]{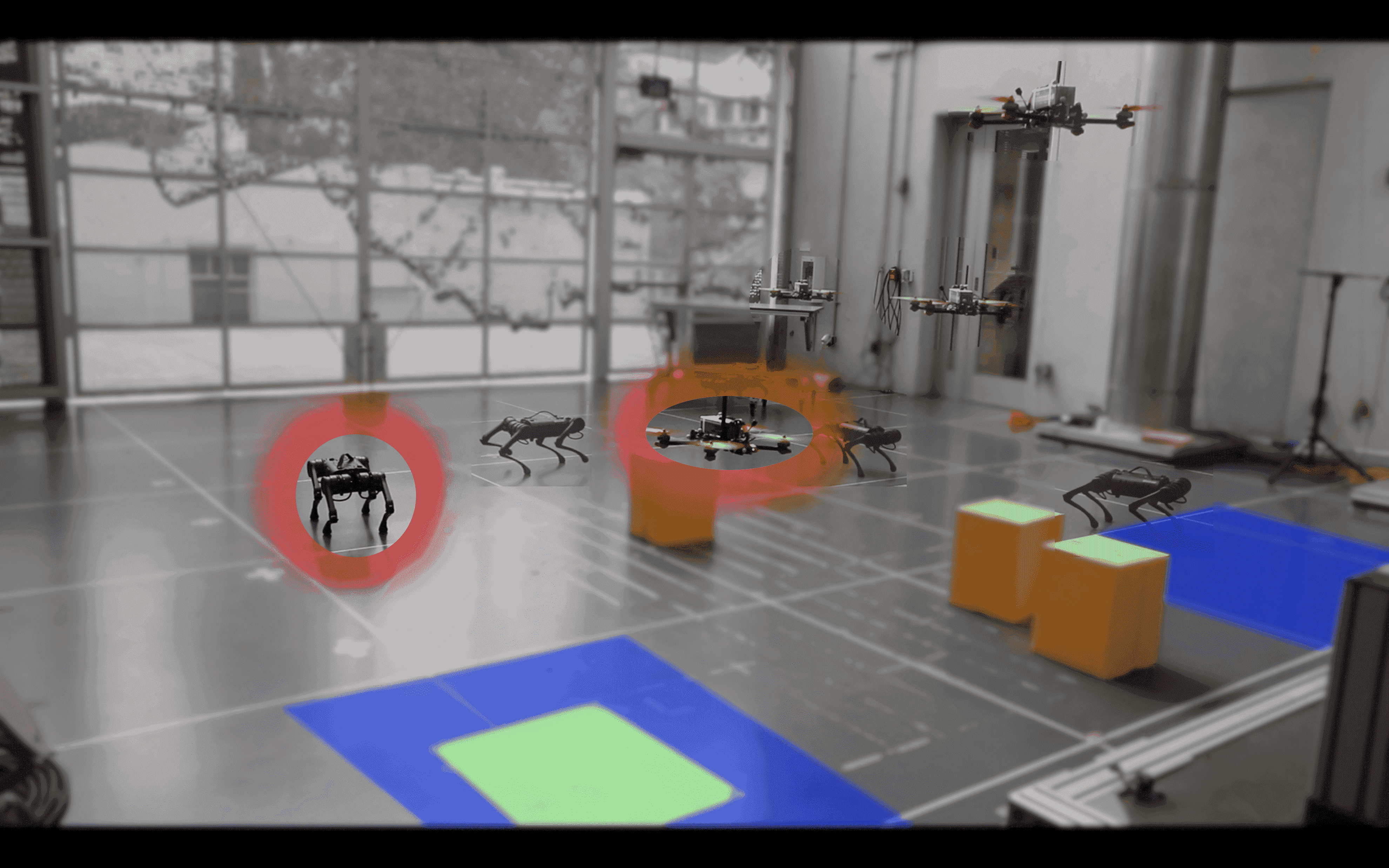}
    \includegraphics[width=0.19\linewidth, trim={15 10 10 10},clip]{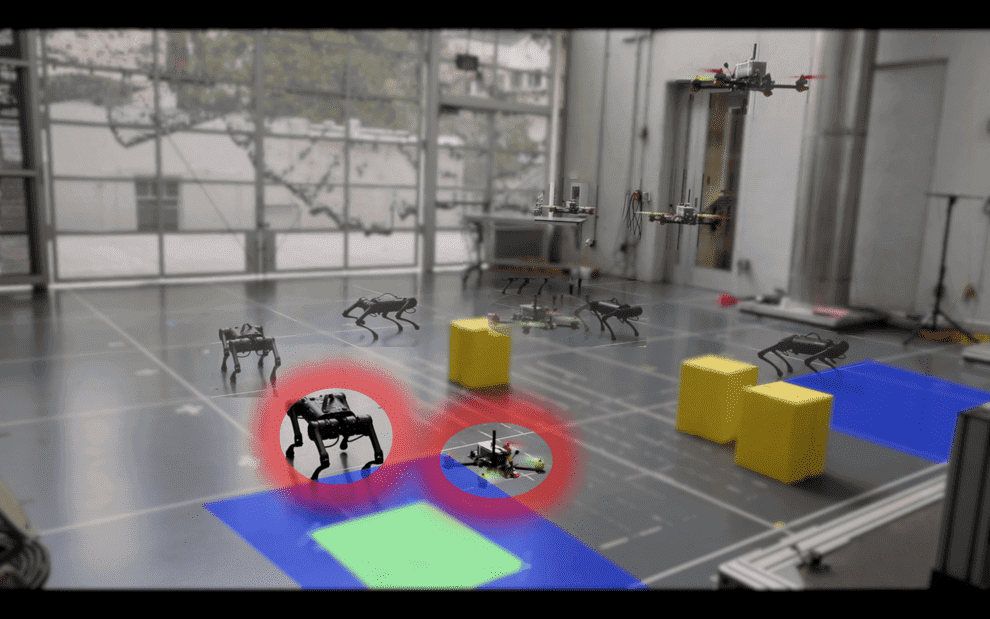}
    \caption{Experimental results from the setup in Figure~\ref{fig:experimentalSetup}. The yellow obstacles in the bottom row move but are always contained inside the static black obstacle region in the top row. Blue and red goal nodes correspond to $e=0$ and $e=1$, respectively. The black dots illustrate an observation node.}
    \label{fig:experimentIllustration}
\end{figure*}

\section{Results} \label{sec:results}
This sections presents the main results: simulations for single and multiple agents, and an implementation on hardware. Costs are quadratic as described in assumption 3 of~\cite{rosolia2021mixedobservable}.\footnote{Code available at https://github.com/kasperjo/MixedObservableRRT.git.}
\subsection{Single-agent simulation and benchmark comparison} We first compare the MORRT to a standard RRT benchmark algorithm on a single-agent example. For the RRT benchmark, the agent commits to a single goal (the most likely one) using a standard RRT, and if the agent crosses an observation area, the belief is updated and the agent recommits to the new best target.

The agent is initialized under an obstacle in a square environment (Figure~\ref{fig:singleAgentExample}). At the bottom there is a hill where the agent can see above the obstacle and receive an $80\%$ accurate observation. On the other side of the obstacle there are two more obstacles; in-between these the agent can view both goal nodes and receive a $90\%$ accurate observation. Finally, at the top, once the agent is near the goal nodes, it receives a perfect observation. The initial belief is $[0.5,0.5]$, meaning that the probabilities of the target being in the top left and top right are both $50\%$.

Figures~\ref{subfig:singleAgent_path1}-\ref{subfig:singleAgent_path8} show the MORRT plan for the eight possible realizations of a mission; observation nodes are illustrated by black dots. The agent first moves down to collect an observation. If the first observation is $o_1=0$ it goes around the obstacle to the left (Figures~\ref{subfig:singleAgent_path1}-\ref{subfig:singleAgent_path4}); otherwise it goes around to the right (Figures~\ref{subfig:singleAgent_path5}-\ref{subfig:singleAgent_path8}). The agent then moves in-between the top obstacles to collect another observation, before moving either left (if $o_2=0$) or right (if $o_2=1$). Finally, at the top observation area the agent receives a perfect observation and commits left or right.

Figure~\ref{subfig:benchmark} shows the plan generated by the standard RRT benchmark algorithm. The agent first commits to one target (the top left in this case) and misses the bottom and middle observation areas. Once at the top left target, the agent receives a perfect observation; if this observation is $o=0$, it stays at the top left; otherwise it must travel to the top right.

Table~\ref{table:bemchmark_comparison} shows a quantitative comparison between the MORRT and the RRT benchmark.
\begin{table}[t]
\footnotesize
\caption{\footnotesize MORRT compared to the standard RRT benchmark.}
\label{table:bemchmark_comparison}
\centering
\begin{tabular}{ccccc} \toprule
 & Expected cost  & Best cost                          &    Worst cost             & Time [s]        \\ \midrule
MORRT & $\textbf{2200}$   & $1560$       & $3980$         & $200$\\
Benchmark & $2890$   & $\textbf{1270}$       & $\textbf{3760}$         & $\textbf{30}$\\
\bottomrule
 \end{tabular}
\end{table}
The realized cost of the best and worst case scenarios are lower for the benchmark. This is because the benchmark only makes a single observation, and takes the shortest path there, while the MORRT visits all observation areas to reduce uncertainty. The expected cost is much lower for the MORRT than the benchmark. Computational time for the MORRT is higher than for the benchmark, as expected, since the benchmark is a standard RRT, while the MORRT finds a tree of RRTs, where the chosen path depends on the observation.\footnote{The computational times can be drastically reduced by, for instance, using C++. The times are just illustrated to show a relative comparison.}

\subsection{Multi-agent simulation} \label{sec:five_agent_example}
Figure~\ref{fig:multi_agent_simulation} illustrates the system. Two quadrupeds and three drones start at $(x,y)=(0,2.5)$. The quadrupeds must avoid the obstacle. There are five observation areas, and the initial belief is $[0.5,0.5]$. We assign the quadrupeds to the areas close to the obstacle, and the drones to the remaining three areas; this reduces computational time as mentioned in Section~\ref{sec:multi_agent_extension}. Observation areas around the goal nodes yield $90\%$ accurate observations, and other observations are $80\%$ accurate. The plan is illustrated in Figure~\ref{fig:multi_agent_simulation}. There are six outcomes, but due to symmetry we show the results for when the agents commit to the top goal node. Observation nodes, where agents make and communicate observations, are illustrated by black dots. The quadrupeds go around the obstacle to collect one observation each. If the observations agree, all robots go to the target (Figure~\ref{subfig:multi_agent_1}). If the observations contradict, two drones make additional observations, one in the top right and one in the middle. If these observations align, the robots commit to the target (Figure~\ref{subfig:multi_agent_2}); otherwise the third drone makes an additional observation in the bottom right before the robots commit (Figure~\ref{subfig:multi_agent_3}).

\subsection{Hardware experiment} \label{sec:experiments}
\subsubsection{Experimental setup and hardware models}
The setup is illustrated in Figure~\ref{fig:experimentalSetup}. A quadruped and a drone are initiated in the origin of a 5m$\times$5m environment, and the mission objective is to locate the science sample. The initial belief is $[0.5,0.5]$ and observations are perfect.

We fix the $z$-coordinate of the drone and compute a plan in $\mathbb{R}^2$. We then model the agents as unicycles~\cite{aicardi1995} and leverage two MPC algorithms to make them follow MORRT way-points. The plan is computed offline, while the MPC algorithms act locally online. We plan for all possible observations, but during the experiment we fix the outcome and feed observations through a centralized information flow. At 60Hz the MPC algorithms solve for velocity and yaw-rate inputs, which are fed to the quadruped and drone.

\subsubsection{Experimental results} 
The experiment is illustrated in Figure~\ref{fig:experimentIllustration}. The quadruped observes $o_1=1$ in the second snapshot, and communicates to the drone; recall that we fix the observation before-hand, but a full plan is computed. Since the observation is perfect, both agents move to the top right goal---taking into account the obstacle for the quadruped---without making another observation. The drone arrives first and waits for the quadruped to arrive.

\section{Conclusion} \label{sec:conslusion}
We introduced the MORRT, which plans a mission for heterogeneous multi-agent exploration, and showed how autonomy can be enabled when multiple agents move and act simultaneously. The approach was illustrated, both in simulation and on hardware. It was found that agents made intelligent collaborative decisions based on their observations from the environment. The problem formulated in this paper can be applied to numerous real-world settings, including search missions, such as the Mars exploration task or rescue operations, where the objective is to locate a hidden target. There are several directions for future work. These include extending the MORRT to account for system dynamics, adding communication constraints to mimic a distributed setting, and analyzing the MORRT convergence properties.

\balance


\renewcommand{\baselinestretch}{0.97}

\bibliographystyle{IEEEtran} 
\bibliography{IEEEabrv,main}

\end{document}